\documentclass[12pt]{article}

\usepackage{lipsum}
\usepackage{ucs}
\usepackage[english]{babel}
\usepackage{graphicx}
\usepackage{amsmath}
\usepackage{authblk}
\usepackage{hyperref}

\usepackage{geometry}
\newgeometry{vmargin={15mm}, hmargin={15mm,15mm}}
\usepackage{indentfirst}
\title{\textbf{AmphibianDetector: adaptive computation for
moving objects detection}}
\author[1,2]{Svitov David}
\author[1]{Alyamkin Sergey}
\affil[1]{Expasoft LLC, \protect\\
11/1 Nikolaeva street, Novosibirsk, 630090, Russia}
\affil[2]{Institute of Automation and Electrometry of the SB RAS, \protect\\
1, Academician Koptyug ave., Novosibirsk, 630090, Russia \protect\\
E-mail: d.svitov@expasoft.tech}
\date{}
\begin{document}

\maketitle

\begin{figure}[h]
\center{\includegraphics[scale=0.45]{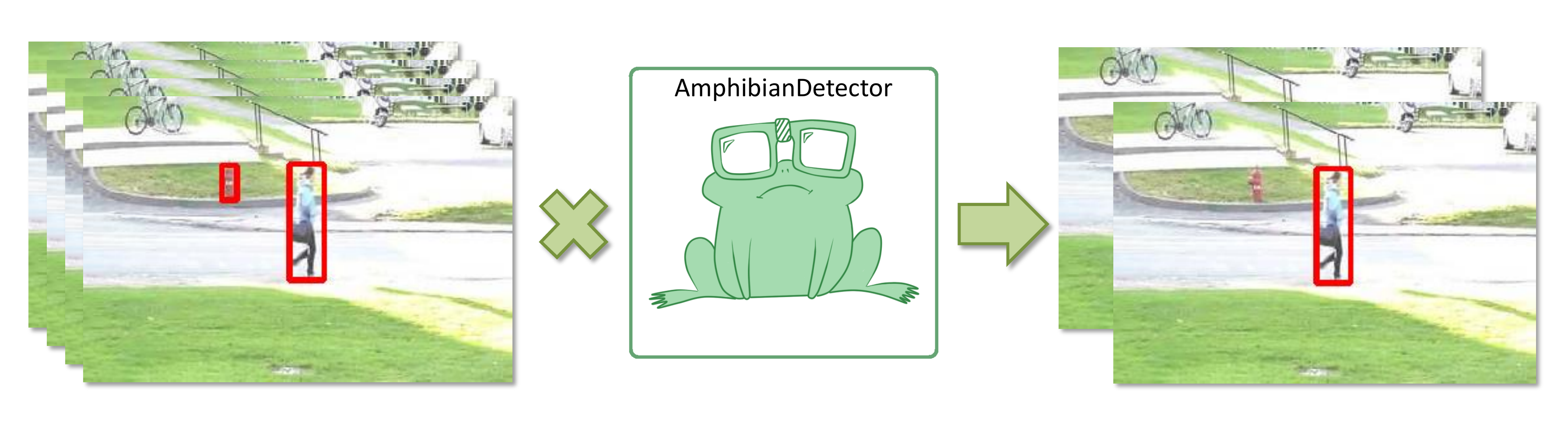}}
\caption{The proposed approach allows effectively detect moving objects of the target class on video. It increases the mean average precision by filtering out false-positive detections of static objects and reduces the average frame processing time due to adaptive computation - the detector partially processes the frame if it does not contain the target moving objects.}
\label{fig:schema}
\end{figure}

\section{Abstract}
Convolutional neural networks (CNN) allow achieving the highest accuracy for the task of object detection in images. Major challenges in further development of object detectors are false-positive detections and high demand of processing power. In this paper, we propose an approach to object detection which makes it possible to reduce the number of false-positive detections by processing only moving objects and reduce the required processing power for algorithm inference.

The proposed approach is a modification of CNN already trained for object detection task. This method can be used to improve the accuracy of an existing system by applying minor changes to the algorithm. The efficiency of the proposed approach was demonstrated on the open dataset "CDNet2014 pedestrian". The implementation of the method proposed in the article is available on the GitHub: \url{https://github.com/david-svitov/AmphibianDetector}

\section{Introduction}

Nowadays, object detectors based on neural networks have established themselves as the most accurate tool for detection of objects in a video stream or on a single frame \cite{girshick2014rich, girshick2015fast, ren2015faster}. Recently, neural network detectors have reached fairly high execution speeds on low-power systems \cite{redmon2016you, redmon2017yolo9000, redmon2018yolov3}. The neural network detectors allow organizing the traffic tracking or an outdoor surveillance system based on a single board computer. Reducing the load on such computer will extend its lifetime. In tasks related to detection, whether it is road traffic or people for an outdoor surveillance system, it seems inappropriate to process the static frames since the state of the environment does not change. In this paper, we propose an approach that allows us to filter out static frames effectively without the use of additional computational resources.

For some systems, the false-positive detections may have a significant negative effect on the work of the entire system. For example, false-positive detections in a face recognition system may lead to undefined behavior when the system tries to process a background fragment as a face. Often in such cases, a motion detector is used and only frames with the moving objects are processed \cite{stauffer1999adaptive, st2014subsense}. Usually, these approaches for a static camera are based on building a background model and calculating the difference between that model and the current frame. In this work, we propose to combine the stage of building a stable background model and detection. To do this, the intermediate layer's features of the detector are used as a background model. This approach does not require additional calculations and allows one to obtain a robust background model.

To solve the problem of filtering out the static frames and filtering false-positive detections, it is not necessary to perform the pixel-wise motion segmentation \cite{babaee2018deep, lim2020learning}. It is enough to understand whether there is movement inside the resulted bounding boxes or not. This allows us to optimize the motion detection procedure by considering a grid of a given scale. The idea of using an intermediate feature map to compare frames on video is close to the idea of perceptual loss \cite{johnson2016perceptual}, widely used for training the generative neural networks \cite{wei2020unsupervised, fritsche2019frequency}.  In such approaches, feature maps from the CNN such as VGG-19 \cite{simonyan2014very} are used to calculate the similarity between the generated and the target image. In this work, we propose to use an intermediate feature map of the detector network that does not increase the computational cost.

The biological inspiration for this work was the vision system of amphibians. When an amphibian is motionless, the amount of visual information it collects is much less than that of mammals. The amphibian sees only moving objects. Similarly, the machine vision model proposed in this work reacts only to moving objects which reduces the computational load. The main contribution of this work is that it proposes to use intermediate feature maps of the detector for frame comparison and thus reduce the load on the calculating device and reduce the number of false positives.

The work consists of three parts: Related Works, Proposed Approach, and Experiments. The first part describes the work on object detection, motion detection, and previous approaches that combine object and motion detection. In the next section, we describe the proposed approach. And in the final section, we show its effectiveness experimentally on an open dataset.

\section{Related Works}
\textbf{Object detection.} Object detection is one of the key tasks of computer vision. The most widely used models are based on RCNN \cite{girshick2014rich, girshick2015fast, ren2015faster}, YOLO \cite{redmon2016you, redmon2017yolo9000, redmon2018yolov3}, and SSD \cite{liu2016ssd}.

\textit{Faster r-cnn.} RCNN-based detection models use the regions proposal algorithm and neural network to classify those regions. S. Ren et al \cite{ren2015faster} proposed the Faster r-cnn approach using the Region Proposal Network. That network uses the same feature maps as the detector which significantly reduces the cost of regions generation.

\textit{You only look once.} J. Redmon et al \cite{redmon2016you}  were the first to propose a single neural network approach to perform detection. YOLO splits the input image into a grid where bounding boxes and class detection probabilities are predicted for each cell. Fully connected layers are used to predict bounding box coordinates and class probabilities. This approach does not include the stage of regions generation and encapsulates the entire logic within a single neural network. This allowed the authors to get a high speed of this architecture. In YOLOv2 \cite{redmon2017yolo9000}, the authors have replaced the fully connected layer with anchor boxes. In YOLOv3 \cite{redmon2018yolov3}, the authors used more accurate neural network architecture for feature extraction and improved the detection of objects of different sizes.

\textit{SSD: Single Shot MultiBox Detector.} Wei Liu et al \cite{liu2016ssd} proposed another approach to object detection in the image using a single neural network. The SSD generates templated bounding boxes with different aspect ratios and different scales for the image feature map. The neural network then predicts the probabilities of finding the target classes in the bounding boxes and refines their coordinates. The main difference of this method from YOLO is the use of feature maps from different layers which increases the accuracy of objects detection of different sizes.

\textbf{Motion detection.} The task of separation of moving objects from the background has been interesting to the community for a long time. The general approach to motion detection in images consists of two main steps. The first step is to build a background model that is a static part of the image. The second step detects motion as the difference between the current frame and the background model. Classic approaches to motion detection used a reference frame as the background model. The Gaussian Mixture Model \cite{stauffer1999adaptive} proposed by Stauffer C. and Grimson W. E. L. became widespread. This approach took into account changes in the background model associated with a dynamic background and changing lighting conditions. St-Charles P. L. et al \cite{st2014subsense} proposed the SuBSENSE approach that uses spatio-temporal binary features as well as color information to detect changes. Wang Y. et al \cite{wang2014cdnet} have proposed an open test dataset for pixel-wise motion detection under various settings.

Deep machine learning methods have made it possible to improve the accuracy of motion detection in a video stream. To the best of our knowledge, M. Braham et al \cite{braham2016deep} were the first to suggest the usage of CNN to build a background model. They were building a grayscale background model based on N image frames. Then, the scene-specific CNN model was trained to segment the moving objects using the background model and the current frame. Babaee M. et al \cite{babaee2018deep} have developed the idea of CNN usage to detect motion in an image. They used SuBSENSE \cite{st2014subsense} and Flux Tensor \cite{wang2014static} to build a background model. Then, feeding the background model and the current frame to the CNN, they performed pixel-wise motion detection. Ang Lim L. and Yalim Keles H. \cite{lim2018foreground} proposed FgSegNet. It is based on triplet CNN using different image scales for the encoder. The usage of different scales allowed them to take into account the context where a fragment is located on one scale according to another. They then used a decoder to perform the pixel-wise motion detection. In their next work, Ang Lim L. and Yalim Keles H. \cite{lim2020learning} proposed FgSegNet-v2. This method extended the Feature Pooling Module of FgSegNet by introducing features fusion inside this module which is capable of extracting multi-scale features within images.

A disadvantage of the existing CNN-based approaches is the need to train an additional neural network. This increases the resources required to build and deploy such system. The usage of additional CNN can negatively affect the frame processing time. In this paper, we propose an approach that does not require additional computing resources for motion detection, since a deep representation of the frame from the intermediate layer of the detector network is used.

\textbf{Combining object and motion detection.} To increase the speed of the object detector, it can be combined with a motion detector. With this approach, frames that do not contain moving objects will not be processed by the object detector. Kang D. et al proposed the Noscope \cite{kang2017noscope} approach. Noscope uses an inference-optimized model search that is optimal for a cascade of difference detectors and specialized networks for a given video, target object, and reference neural network. The use of the Mean Square Error (MSE) between the reference frame and the current frame as a difference detector made it possible to obtain a high speed of video stream processing. The accuracy of the resulting cascade is the same as the accuracy of the reference neural network. However, this approach has several disadvantages. First, it requires retraining the system for each camera and scene. Secondly, motion detection based on reference frame comparison is not robust to the dynamic background and changing lighting.

Another approach to combine motion detector and object detector was proposed by Yu R., Wang H., and Davis L. S in the ReMotENet \cite{yu2018remotenet}. They targeted to detect the movement of the relevant object on video. To do this, they analyzed the entire video using three-dimensional (3D) convolutions. The disadvantage of this approach is the need to process the entire video which limits its applicability for outdoor surveillance cameras. Also, ReMotENet only answers the question of whether there was the movement of the relevant object on video, but does not localize it.

We propose an approach that uses a feature map obtained from the intermediate layers of the detector's neural network. This allows us to compare frames based on deep features. If there is no movement, the processing of the frame by the network stops at this stage.

\section{Proposed Approach}
\begin{figure}[h]
\center{\includegraphics[scale=0.45]{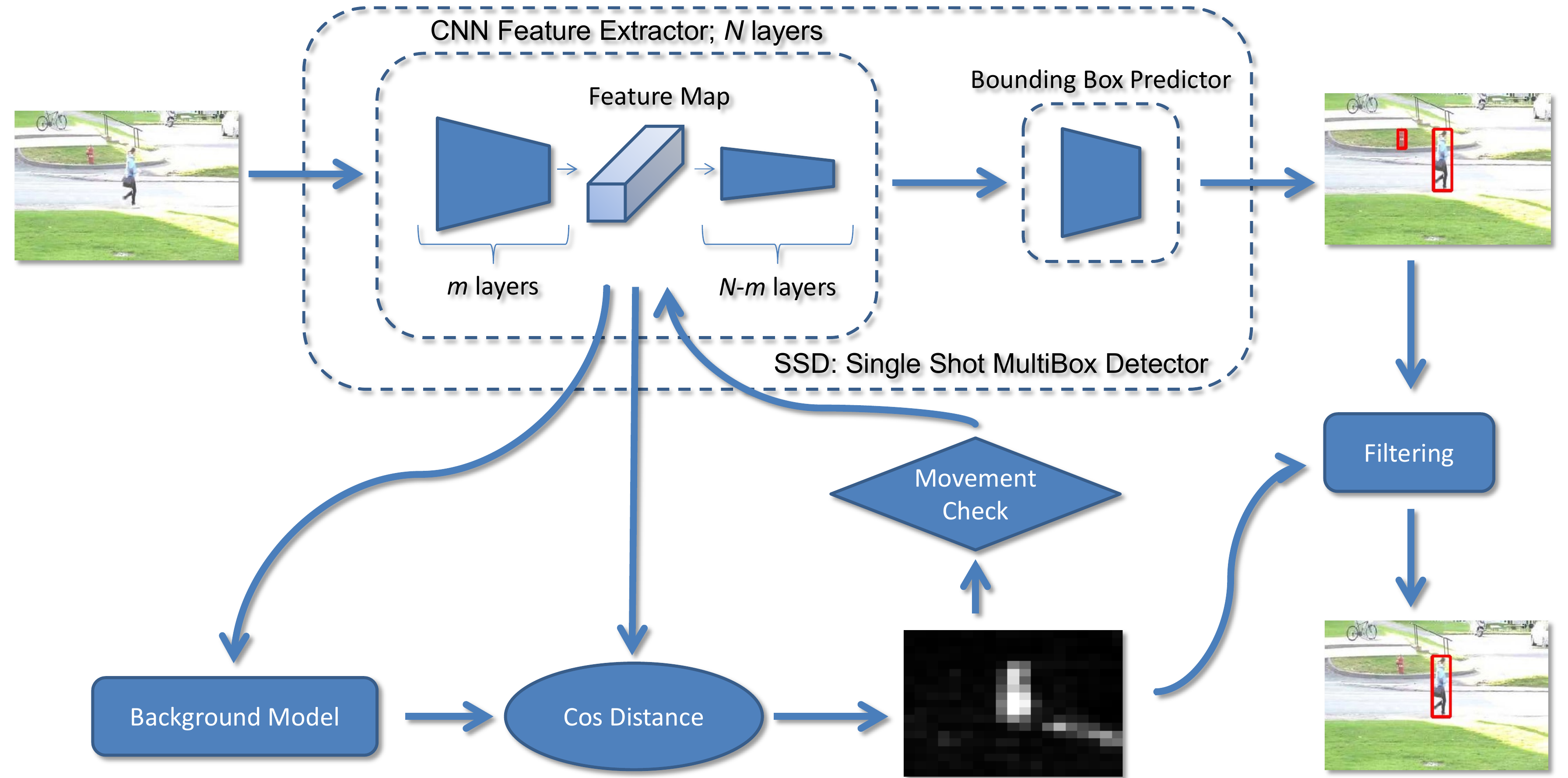}}
\caption{The input image is processed by the first m layers of the feature extractor which is part of the object detection model. The feature maps obtained at this stage are used to build the background model. The feature map for the current frame is compared with the background model by cosine similarity. The resulting motion map is used to check whether it is necessary to end the inference of the detector. The motion map is also used to filter the bounding boxes in order to filter out false-positive detections corresponding to static objects.}
\label{fig:schema}
\end{figure}

In this paper, we propose an approach to the detection of moving objects in a video stream (Fig. \ref{fig:schema}). This method allows us to reduce the average processing time of a video frame and reduce the number of false-positive detections. The key idea of this method is to use the intermediate feature maps from the detector feature extractor to determine the regions of motion in the frame. In this case, the feature map of the intermediate layer is considered as a matrix of vectors where each vector describes the corresponding area in the input image. Then, by comparing these vectors for the current frame with the vectors representing the background model, we estimate changes in different parts of the image.

Let \begin{math}F_1..F_i..\end{math} denote the sequence of frames in the video stream. Here \begin{math}F_i\end{math} denotes the \begin{math}i\end{math}-th frame. We use SSD \cite{liu2016ssd} for object detection as it is the most common neural network architecture for this task. The network architecture for object detection consists of a feature extractor and a part of the network predicting the coordinates of the bounding boxes. The feature extractor is a convolutional neural network such as MobileNetV2 \cite{sandler2018mobilenetv2} or Inception \cite{szegedy2015going}. It consists of \begin{math}N\end{math} convolutional layers \begin{math}L_1..L_N\end{math}. To obtain a feature map from an intermediate layer, value \begin{math}m \in [1..N]\end{math} is selected. We propose to use the output of the \begin{math}L_m\end{math} layer as the feature map. The feature map for the frame \begin{math}i\end{math}  will be denoted by \begin{math}M_i\end{math}. The resulting feature map has the size \begin{math}H_m \times W_m \times C_m\end{math}. Here \begin{math}H_m\end{math} and \begin{math}W_m\end{math} correspond to the height and width of the feature map. \begin{math}C_m\end{math} is the number of channels and can be interpreted as the dimension of vector at each point of the feature map.
The choice of \begin{math}m\end{math} determines which area of the input image will correspond to each vector on the feature map, and also how robust the vector representation is for the artefacts of the input image. By adjusting the \begin{math}m\end{math} parameter, you can achieve the desired ratio between the speed and accuracy of the motion detector. The choice of the optimal \begin{math}m\end{math} will be discussed in the "Experiments" section.

The size of the area of input image corresponding to one vector on the feature map for a given \begin{math}m\end{math} is calculated as follows. Let \begin{math}H_{in} \times W_{in} \times 3\end{math} be the size of the three-channel input image, where \begin{math}H_{in}\end{math} and \begin{math}W_{in}\end{math} are respectively the height and width of the input image. Then, for a given \begin{math}m\end{math}, the size of the input area will be calculated as \begin{math}H_{in}/H_m \times W_{in}/W_m\end{math}.

To detect changes in the \begin{math}F_i\end{math} frame, its corresponding feature map \begin{math}M_i\end{math} is compared with the background model \begin{math}M_{bg}\end{math}. The background model is initialized with some reference frame \begin{math}M_{init}\end{math}. The reference frame must not contain the moving objects, since their displacement will cause false-positive detections.

To compare the background model \begin{math}M_{bg}\end{math} and the feature map of the frame \begin{math}M_i\end{math}, the cosine similarity between their vector representations is calculated. For each grid cell, corresponding to the layer \begin{math} m \end{math}, with coordinates \begin{math} x \in [0..W_m], y \in [0..H_m] \end{math} , cosine similarity is calculated as:
\begin{equation}
Diff_i[x, y] = 1 - \frac{M_{bg}[x, y] \times M_i[x, y]}{||M_{bg}[x, y]|| * ||M_i[x, y]||},
\end{equation}
where \begin{math} \times \end{math} is the dot product of vectors, and \begin{math} * \end{math} is the product of the norms of vectors. The closer the value of the motion map \begin{math}Diff_i[x, y]\end{math} is to \begin{math}1\end{math}, the stronger the change in the area of the input image \begin{math}F_i\end{math} is. As Wang H. et al. \cite{wang2018cosface} noted, the usage of cosine similarity to compare vector representations trained using softmax improves consistency compared to Euclidean distance.

Optimization of the detector's inference speed is performed by stopping the processing of frames where no motion is detected. To reject static frames, the maximum value \begin{math}max_{x, y}(Diff_i)\end{math} is compared with the threshold \begin{math}\lambda\end{math}. Frames \begin{math}F_i\end{math} where \begin{math}max_{x, y}(Diff_i)\end{math}  is less than the threshold \begin{math}\lambda\end{math} are used to update the background model.

As noted by M. Babaee et al \cite{babaee2018deep}, the dynamic update of the background model allows adapting it to changes in scene illumination and weather conditions. In our approach, \begin{math}M_{bg}\end{math} is updated based on \begin{math}M_i\end{math} using the following formula:
\begin{equation}
M_{bg} =
 \begin{cases}
   M_{bg}*\alpha + M_i * (1-\alpha) &\text{if $max_{x, y}(Diff_i) < \lambda$}\\
   M_{bg} &\text{otherwise}
 \end{cases}
\end{equation}
By adjusting the \begin{math}\alpha\end{math} parameter, we can define how long the object needs to remain motionless in the frame to become a part of the background model. For \begin{math}\alpha=0\end{math} each frame will be compared with the feature map of the previous frame. The case \begin{math}\alpha=1\end{math} corresponds to comparison of the feature map of the \begin{math}i\end{math}-th frame with the reference one.
For frames with moving objects, after predicting the bounding boxes, additional filtering is performed based on the \begin{math}Diff_i\end{math}. Bounding box coordinates are predicted in normalized values from \begin{math}0..1\end{math} and then scaled to the size of the input image \begin{math}F_i: H_{in} \times W_{in}\end{math} and the motion map \begin{math}Diff_i: H_m \times W_m\end{math}. We denote the bounding boxes for the frame \begin{math}F_i\end{math} by \begin{math}B_{i1}..B_{iM}\end{math}. Each bounding box \begin{math} B_ {ij} \end{math} is specified through its coordinates \begin{math} [left, top, right, bottom] \end{math}, which correspond to the boundary positions \begin{math} x \end{math} and \begin{math} y \end{math} for the given box on the  layer \begin{math} m \end{math}. Then, for the bounding box with the index \begin{math}k\end{math}, the coordinates corresponding to the area in the input image will be calculated as:
\begin{equation}
Bin_{ik}=[B_{ik}.left*W_{in}, B_{ik}.top*H_{in}, B_{ik}.right*W_{in}, B_{ik}.bottom*H_{in}]
\end{equation}
For the motion map \begin{math}Diff_i\end{math} respectively:
\begin{equation}
Bdiff_{ik}=[B_{ik}.left*W_m, B_{ik}.top*H_m, B_{ik}.right*W_m, B_{ik}.bottom*H_m]
\end{equation}
\begin{math}Bin_{ik}\end{math} is removed from the bounding box list if \begin{math}mean_{x, y}(Diff_i[Bdiff_{ik}])\end{math} is less than the threshold \begin{math}\lambda\end{math}.

The proposed method allows us to accelerate the processing of video by stopping the processing of static frames. It also allowed to reduce the number of false-positive detections of the detector for tasks where it is necessary to detect the moving objects.

\section{Experiments}
\begin{figure}[h!]
\center{\includegraphics[scale=0.7]{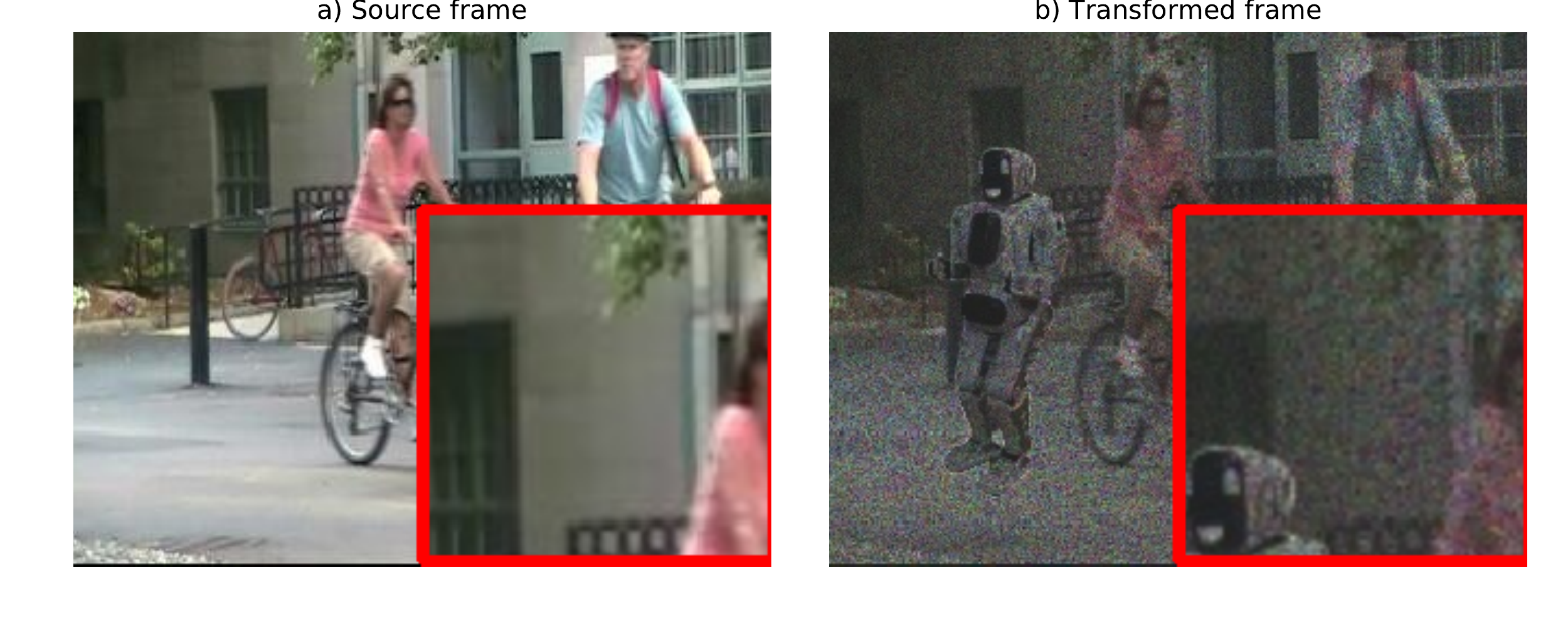}}
\caption{a) An example of frame from the source video; b) A frame from the video after adding a static object looking like a human and multiplying by Gaussian noise ($\mu=0.8, \sigma=0.2$).}
\label{fig:image_compare}
\end{figure}

The proposed AmphibianDetecor approach was compared with the baseline approach to solve the problem of filtering out false-positive detections of static objects. The baseline approach is applied to compare the frames pixels intensity using the L2 distance to determine the motion. To compute the background model, the baseline approach uses selecting the anchor frame and updating the model with an \begin{math} \alpha \end{math} factor as described in the previous section. In our experiments, we evaluated the speed and accuracy of the proposed algorithm. We measured the average frame processing time for all videos in the dataset to evaluate the speed. Precision was measured as the mean average precision (MAP) metric \cite{everingham2010pascal}.  The pedestrian subset of the CDNet2014 dataset \cite{wang2014cdnet} was considered as the dataset for evaluating the algorithm. This subset contains 10 videos of pedestrians in different lighting and weather conditions. The subset contains indoor and outdoor videos. A total of 26248 frames, an average of 2624 frames per video.

\begin{figure}[!h]
\center{\includegraphics[scale=0.6]{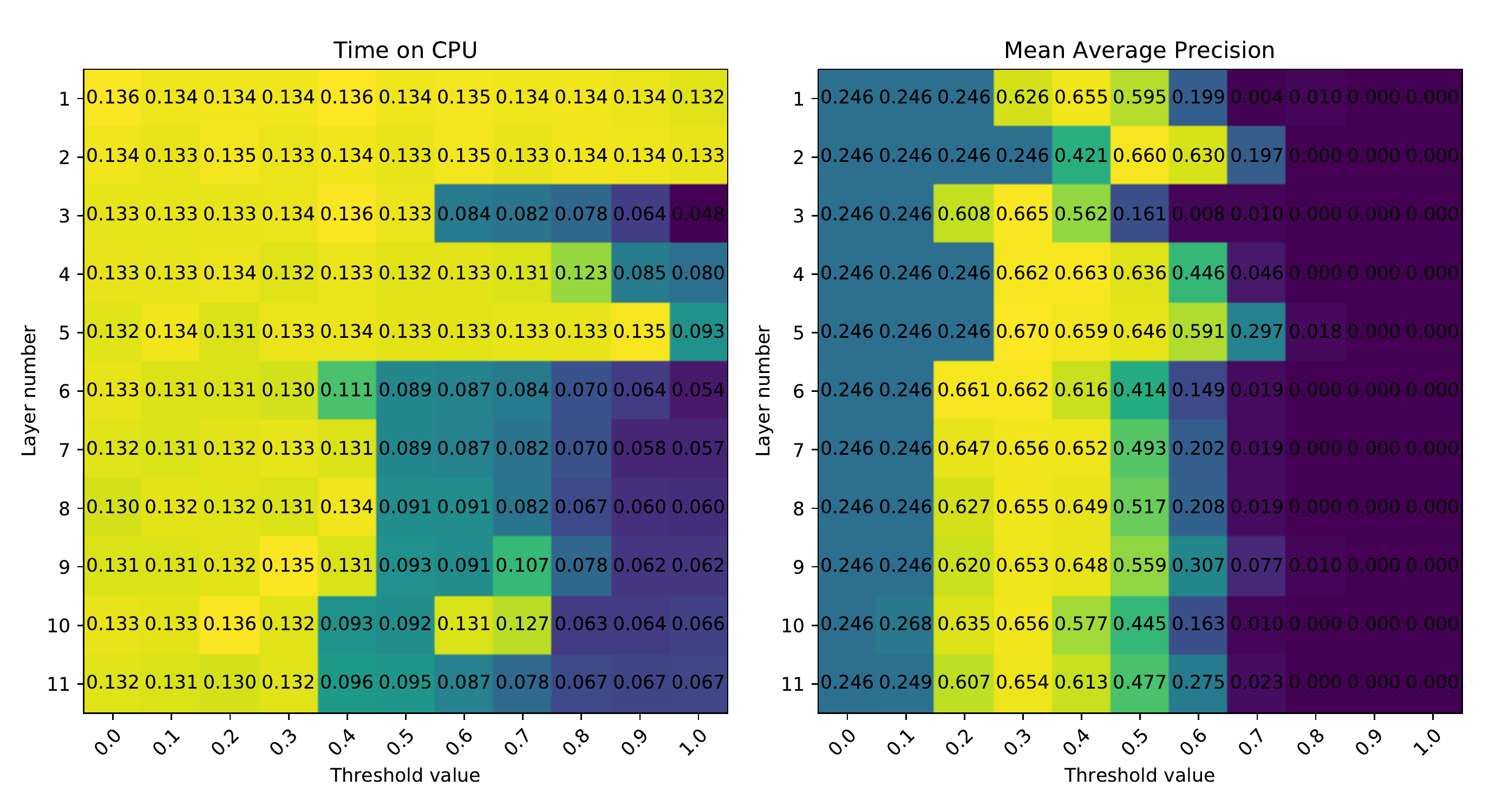}}
\caption{Comparison of MAP and average frame processing time on Intel Core i5-4210U CPU 1.70GHz $\times$ 4 for different values of the layer number $m$ and different threshold values $\lambda$.}
\label{fig:layers_map}
\end{figure}

\begin{figure}[!h]
\center{\includegraphics[scale=0.7]{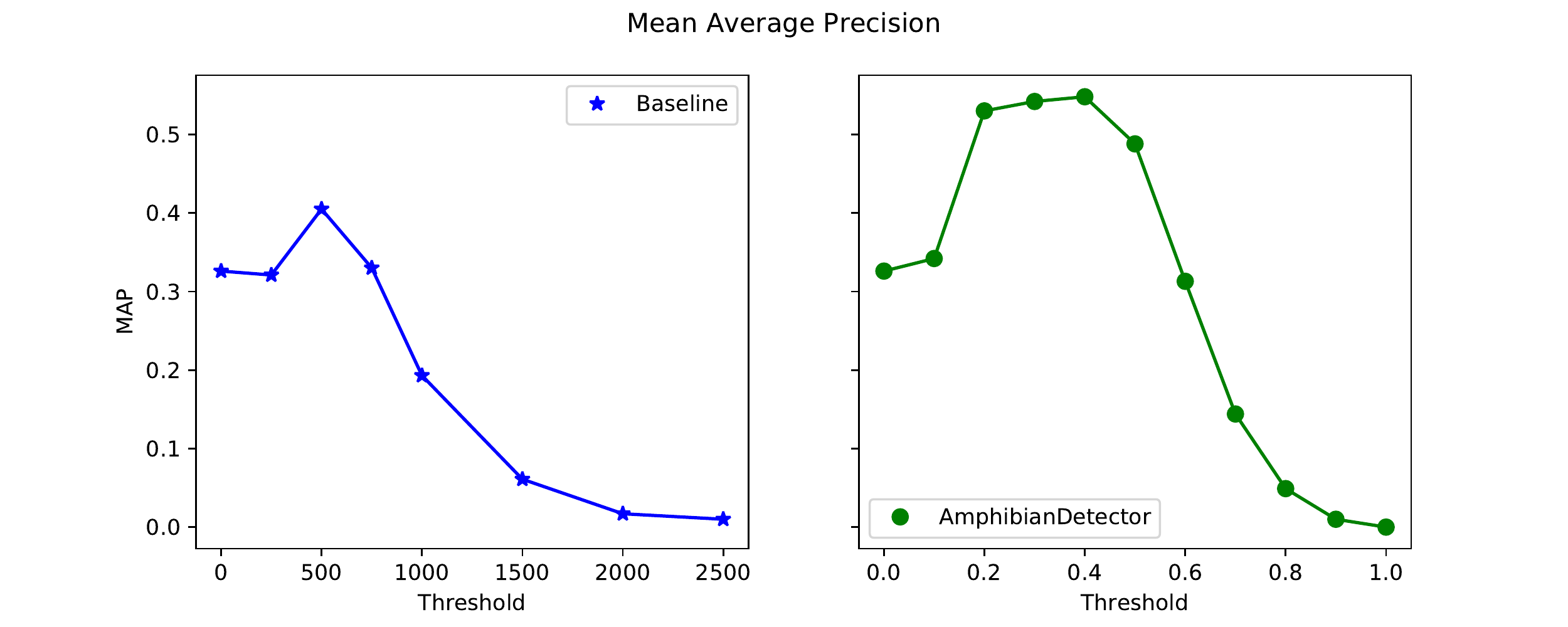}}
\caption{Comparison of the MAP metric for the AmphibianDetector and the baseline approach based on SSD + MobileNetV2. The maximum metric value for the baseline is reached at the threshold $\lambda = 500$. The maximum MAP value using the AmphibianDetector based on SSD + MobileNetV2 is reached at $\lambda = 0.4$.}
\label{fig:map_compare}
\end{figure}

We have simulated cases where the detector performs a false-positive detection on a static object which looks like the object of the target class. To do this, we added a static image of an object resembling a human to the frames of the original video (Fig \ref{fig:image_compare}). Also, videos received from the outdoor cameras are influenced by noise due to the quality of capturing device and video transmission channel to the server. In order to bring the dataset closer to the real case, each frame was modified by adding the noise. The original frames were multiplied by Gaussian noise with \begin{math}\mu=0.8\end{math} and \begin{math}\sigma=0.2\end{math}, simulating the artefacts of the video capturing device.

To compare the AmphibianDetector with the basic approach, we conducted experiments on the SSD \cite{liu2016ssd} architecture with the MobileNetV2 \cite{sandler2018mobilenetv2} feature extractor with an input of \begin{math}300 \times 300 \end{math}. The choice of this architecture was made due to its popularity for embedded systems. The experiments used the network trained on the COCO \cite{lin2014microsoft} dataset. The values of the layer number \begin{math}m\end{math} and the threshold \begin{math}\lambda\end{math} were selected by grid search on one video from "CDNet2014 pedestrian" consisting of 2000 frames. The \begin{math}\lambda\end{math} threshold ranged from 0 to 1, as these are the minimum and maximum values for cosine similarity. The layer number \begin{math}m\end{math} varied between 1 and 11 layers of the feature extractor. From the obtained heat maps (Fig \ref{fig:layers_map}) we see that the highest MAP value equal to 0.67 is achieved when \begin{math}m=5\end{math} and \begin{math}\lambda=0.3\end{math}. And the best ratio of MAP to the average processing time of a frame is achieved with the parameters \begin{math}m=11\end{math} and \begin{math}\lambda=0.4\end{math}.

\begin{figure}[!h]
\center{\includegraphics[scale=0.7]{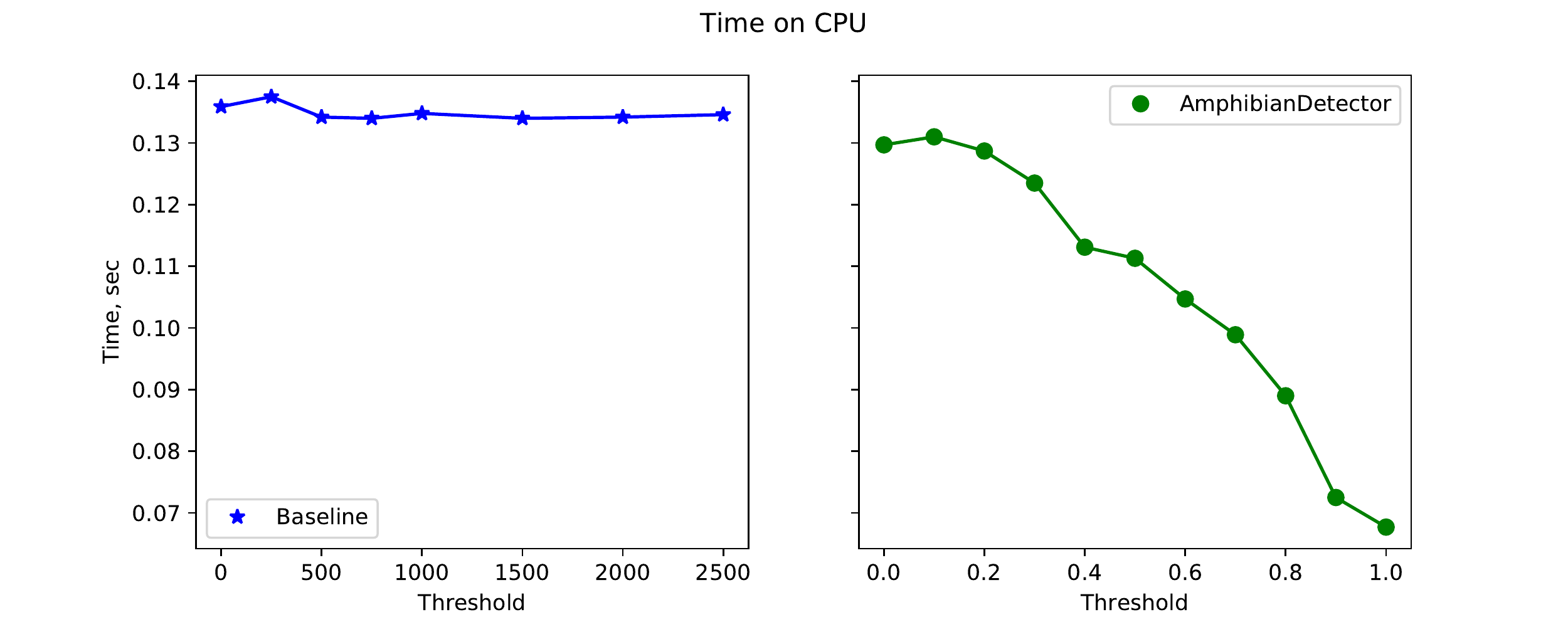}}
\caption{Comparison of average frame processing time for "CDNet2014 pedestrian" on Intel Core i5-4210U CPU 1.70GHz $\times$ 4 for baseline approach and AmphibianDetector based on SSD + MobileNetV2.}
\label{fig:time_compare}
\end{figure}

For the baseline solution, the threshold value \begin{math}\lambda\end{math} varied from 0 to 2500 with the step of 500, since for a larger threshold value the MAP decreased to 0. The AmphibianDetector measurements with the MobileNetV2 + SSD were carried out for the layer number \begin{math}m=11\end{math} showing the best MAP to time ratio for a subset of 2000 frames. The obtained plots show that the MAP for the AmphibianDetector is almost 15\% higher than the MAP for the baseline solution (Fig. \ref{fig:map_compare}) when choosing the best threshold for each approach. The baseline solution reaches 0.405 MAP for the threshold of 500. The AmphibianDetector reaches 0.548 for the threshold of 0.4. It should be noted that in the above experiments, the maximum achievable MAP is limited by the SSD + MobileNetv2 accuracy.

The proposed AmphibianDetector approach reduces the average frame processing time. To compare the average frame processing time, we measured the time on the Intel Core i5-4210U CPU 1.70GHz $\times$ 4. The threshold value \begin{math}\lambda\end{math} varied in the same ranges as for the MAP measurements. It can be seen from the plots (Fig. \ref{fig:time_compare}) that the AmphibianDetector allowed us to reduce the average time for frame processing due to effective filtering out the frames without movement. This is achieved by using feature maps from the network intermediate layer to compare frames.

\begin{table}[h!]
\begin{center}
\caption{Comparison of MAP for SSD with the MobileNetV2 feature extractor and the AmphibianDetector based on SSD + MobileNetV2.}
\begin{tabular}{|l|l|}
\hline
Model                                      & MAP   \\ \hline
SSD+MobileNetV2                            & 0.326 \\ \hline
AmphibianDetector (MNV2; m=11, $\lambda$=0.4) & 0.548 \\ \hline
\end{tabular}
\label{table_maps}
\end{center}
\end{table}

We compared the accuracy of the AmphibianDetector based on SSD + MobileNetV2 with the baseline solution based on SSD + MobileNetV2. The feature map values were taken from the \begin{math}m=11\end{math} layer. The threshold value \begin{math}\lambda\end{math} was set to 0.4. To compare the accuracy, we compared the MAP value for the standard use of detector and AmphibianDetector. SSD + MobileNetV2 enabled to achieve the MAP value equal to 0.326, and the usage of AmphibianDetector allowed us to increase the MAP up to 0.548.

\section{Conclusion}
We have proposed an approach that allows to increase the speed of the detector inference by filtering out frames from the video stream that do not contain moving objects. We have shown experimentally that the proposed approach reduces the average frame processing time compared to the baseline approach based on pixel-level frame comparison. The proposed approach reduces the number of false-positive detections which can be critical in a number of practical applications, for example, face detection for a face recognition system. The source code with the implementation of the proposed approach and experiments is available on the GitHub.

\bibliographystyle{ieeetr}
\bibliography{amphibiandetector}

\end{document}